\documentclass[letterpaper, 10 pt, conference]{ieeeconf}
\IEEEoverridecommandlockouts
\usepackage{cite}
\usepackage{amsmath,amssymb,amsfonts}
\usepackage{graphicx}
\usepackage{textcomp}
\usepackage{adjustbox}
\usepackage{balance}
\usepackage{threeparttable}
\usepackage{soul}
\usepackage{verbatim}
\usepackage{url}
\usepackage{amssymb}
\usepackage{mathtools}
\usepackage{xparse}
\usepackage{textcomp}
\usepackage{comment}
\usepackage{multirow}
\usepackage{array}
\usepackage{nicefrac} 
\usepackage{algpseudocode} 
\usepackage{accents}
\usepackage{xcolor}

\newcommand{\largedot}[2][1.5]{
\accentset{\scalebox{#1}{$\mbox{.}$}}{#2}
}

\makeatletter
\newcommand{\raisemath}[1]{\mathpalette{\raiseMath{#1}}}%
\newcommand{\raiseMath}[3]{\raisebox{#1}[0pt][0pt]{$#2#3$}}
\makeatother

\NewDocumentCommand{\qbar}{O{0.5pt} O{-6.55pt}}{
	\ensuremath{\mathrlap{\raisemath{#2}{\hspace*{#1}{\mathchar'26\mkern-9mu}}} q}%
}

\NewDocumentCommand{\qbars}{O{0.5pt} O{-4.65pt}}{
	\ensuremath{\mathrlap{\raisemath{#2}{\hspace*{#1}{\mathchar'26\mkern-9mu}}} q}%
}

\NewDocumentCommand{\qbarc}{O{0.5pt} O{-5.2pt}}{
	\ensuremath{\mathrlap{\raisemath{#2}{\hspace*{#1}{\mathchar'26\mkern-9mu}}} q}%
}

\NewDocumentCommand{\pbar}{O{-1.5pt} O{-6.65pt}}{
	\ensuremath{\mathrlap{\raisemath{#2}{\hspace*{#1}{\mathchar'26\mkern-9mu}}} p}%
}

\newcommand{\bs}[1]{\boldsymbol{#1}}  
\newcommand{\ts}[1]{\text{#1}}

\renewcommand{\baselinestretch}{0.935}  

\def\BibTeX{{\rm B\kern-.05em{\sc i\kern-.025em b}\kern-.08em
    T\kern-.1667em\lower.7ex\hbox{E}\kern-.125emX}}
\begin{document}

\title{\LARGE \bf 
A New Type of Axis--Angle Attitude Control Law for \\ Rotational Systems: Synthesis, Analysis, and Experiments \\

\thanks{The research presented in this paper was supported by the Joint Center for Aerospace Technology Innovation (JCATI) through \mbox{Award\,$172$}, the Washington State University (WSU) Foundation and the Palouse Club through a Cougar Cage Award to \mbox{N.\,O.\,P\'erez-Arancibia}, and the WSU Voiland College of Engineering and Architecture through a \mbox{start-up} fund to \mbox{N.\,O.\,P\'erez-Arancibia}.}
\thanks{F.\,M.\,F.\,R.\,Gon\c{c}alves and N.\,O.\,P\'erez-Arancibia are with the School of Mechanical and Materials Engineering, Washington State University (WSU), Pullman, WA 99164-2920, USA. R.\,M.\,Bena is with the Department of Mechanical and Civil Engineering, California Institute of Technology, Pasadena, CA 91125-2100, USA. Corresponding authors' email: {\tt francisco.goncalves@wsu.edu} (F.\,M.\,F.\,R.\,G.); {\tt n.perezarancibia@wsu.edu} (N.\,O.\,P.-A.).}%
}

\author{\mbox{Francisco M. F. R. Gon\c{c}alves}, Ryan M. Bena, and \mbox{N\'estor O. P\'erez-Arancibia}}

\maketitle
\thispagestyle{empty}
\pagestyle{empty}

\begin{abstract}
Over the past few decades, continuous quaternion-based attitude control has been proven highly effective for driving rotational systems that can be modeled as rigid bodies, such as satellites and drones. However, methods rooted in this approach do not enforce the existence of a unique \textit{\mbox{closed-loop}} (CL) equilibrium \textit{\mbox{attitude-error} quaternion} (AEQ); and, for rotational errors about the \mbox{attitude-error} Euler axis larger than $\bs{\pi}$\,rad, their \mbox{proportional-control} effect diminishes as the system state moves away from the stable equilibrium of the CL rotational dynamics. In this paper, we introduce a new type of attitude control law that more effectively leverages the \mbox{attitude-error} Euler \mbox{axis--angle} information to guarantee a unique CL equilibrium AEQ and to provide greater flexibility in the use of \mbox{proportional-control} efforts. Furthermore, using two different control laws as examples---through the construction of a strict Lyapunov function for the CL dynamics---we demonstrate that the resulting unique equilibrium of the CL rotational system can be enforced to be uniformly asymptotically stable. To assess and demonstrate the functionality and performance of the proposed approach, we performed numerical simulations and executed dozens of \mbox{real-time} \mbox{tumble-recovery} maneuvers using a small quadrotor. These simulations and flight tests compellingly demonstrate that the proposed \mbox{axis--angle-based} method achieves superior flight performance---compared with that obtained using a \mbox{high-performance} \mbox{quaternion-based} controller---in terms of stabilization time.
\end{abstract}

\section{Introduction} 
\vspace{-1ex}
\label{Section01}
The specialized literature has reported extensive research on attitude control methods for rotational systems that can be modeled as rigid bodies, including satellites, drones, and microswimmers\mbox{\cite{GoncalvesFMFR2024I,GoncalvesFMFR2024II,GoncalvesFMFR2024III,BenaRM2022,BenaRM2023I,MayhewCG2009,MayhewCG2011I,MayhewCG2011II,PratamaB2018,MokhtariA2004,KangCW2011,LeeT2011,LeeT2018,WuG2014,WeiJ2017,ThunbergJ2014,BlankenshipEK2024, TrygstadCK2025I, TrygstadCK2025II}}. In this context, we can distinguish three main types of prevalent controllers: \mbox{quaternion-based}\mbox{\cite{GoncalvesFMFR2024I,GoncalvesFMFR2024II,GoncalvesFMFR2024III,BenaRM2022,BenaRM2023I,MayhewCG2009,MayhewCG2011I,MayhewCG2011II}}, \mbox{Euler-angle--based}\mbox{\cite{PratamaB2018,MokhtariA2004,KangCW2011}}, and rotation-matrix--based (geometric)\mbox{\cite{LeeT2011,LeeT2018,WuG2014}}. By contrast, the published research on attitude control directly based on the Euler \mbox{axis--angle} representation---despite its multiple advantages---is scarce, although the relevant information required to adopt this approach is contained in the rotation matrices and quaternions employed by other methods. For example, in the formulation of \mbox{quaternion-based} attitude control laws, typically, a proportional torque term is defined by simply multiplying the vector part of the instantaneous \textit{\mbox{attitude-error} quaternion} (AEQ) by a \mbox{positive-definite} matrix, which therefore includes a \mbox{$\sin\frac{\Theta_{\ts{e}}}{2}$} factor, in which $\Theta_{\ts{e}}$ is the instantaneous rotation error about the \mbox{attitude-error} Euler axis. It appears that the only rationale behind this design choice is to simplify the mathematical analysis of the \textit{\mbox{closed-loop}} (CL) system dynamics. Additionally, in most reported applications, the proportional term in the \mbox{quaternion-based} law is modified through a switching condition, included to avoid unwinding behavior by forcing the controller to apply the proportional torque effort in the direction of the shortest rotational trajectory required to eliminate the attitude error\mbox{\cite{BenaRM2022,BenaRM2023I,MayhewCG2009,MayhewCG2011I,MayhewCG2011II}}. Simple analyses indicate that geometric controllers also apply torque in this direction of shortest rotation, which---as experimentally shown~in\mbox{\cite{GoncalvesFMFR2024I,GoncalvesFMFR2024II,GoncalvesFMFR2024III}}---is not necessarily the best choice for every kinematic situation.

In this paper, we present a new type of \mbox{axis--angle} law to control the attitude of \mbox{rigid-body} rotational systems, which guarantees the existence of a unique CL fixed AEQ and does not diminish the \mbox{proportional-control} effort as the system state moves away from the stable CL fixed AEQ---quantified as a rotation about the Euler axis determined by the instantaneous CL AEQ. This research was prompted by phenomena observed during the development and implementation of \mbox{quaternion-based} switching schemes conceived to account for both the direction and magnitude of the flier's angular velocity---in order to heuristically minimize a figure of merit---when choosing the direction in which the proportional torque control effort is applied\mbox{\cite{GoncalvesFMFR2024I,GoncalvesFMFR2024II,GoncalvesFMFR2024III}}. As a result, these controllers often exert the proportional torque in the direction of the longest rotational trajectory required to eliminate the attitude error---i.e., rotations larger than \mbox{$\pi$\,rad}---which highlights the potential practical utility of this proposed control approach when used in combination with those types of schemes. Furthermore, using two different control laws as examples---through the construction of a strict Lyapunov function for the CL dynamics---we demonstrate that the resulting unique equilibrium of the CL rotational system can be enforced to be uniformly asymptotically stable by satisfying a \mbox{positive-definite} condition. To test and demonstrate the suitability of the proposed approach, we implemented numerical simulations and performed dozens of \mbox{real-time} \mbox{tumble-recovery} maneuvers using a \mbox{$32$-g} quadrotor. The obtained simulation and experimental data provide compelling evidence of the high performance characteristic---in terms of stabilization time---of controllers synthesized using attitude laws of the proposed type. 

The rest of the paper is organized as follows. Section\,\ref{Section02} provides an overview of the \mbox{open-loop} attitude dynamics of a generic \mbox{rigid-body} rotational system, and describes the synthesis of the \mbox{quaternion-based} controller used as the starting point of the presented research and benchmark. Section\,\ref{Section03} describes the proposed \mbox{axis--angle} attitude control approach, presents the synthesis of two controllers, and discusses the stability of the fixed points associated with the two resulting CL dynamics. Section\,\ref{Section04} presents numerical simulations and experimental results---obtained using a small quadrotor---that demonstrate the functionality and high performance---measured in terms of stabilization time---of the introduced methodology. Last, Section\,\ref{Section05} summarizes the results presented in the paper, discusses some conclusions, and states a direction for future research.

\vspace{1ex}
\textit{\textbf{Notation--}}
\begin{enumerate}
\item Italic lowercase and Greek symbols represent scalars, e.g., $p$ and $\Theta$; bold lowercase symbols represent vectors, e.g., $\bs{p}$; bold uppercase symbols represent matrices, e.g., $\bs{P}$; and, bold crossed lowercase symbols represent quaternions, \mbox{e.g., $\bs{\pbar}$}.
\item The sets of integers, nonnegative integers, and positive integers are denoted by $\mathbb{Z}$, $\mathbb{Z}_{\geq 0}$, and $\mathbb{Z}_{> 0}$, respectively. Similarly, the sets of reals, nonnegative reals, and positive reals are denoted by $\mathbb{R}$, $\mathbb{R}_{\geq 0}$, and $\mathbb{R}_{> 0}$, respectively.
\item The set of unit quaternions is denoted by $\mathcal{S}^3$. The special orthogonal group in the \mbox{three-dimensional} space is denoted by $\mathcal{SO}(3)$.
\item The symbol $\times$ denotes the \mbox{cross-product} between two vectors. Multiplication between two quaternions is denoted by $\otimes$.
\item The symbols $\succ$, $\prec$, $\succeq$, and $\preceq$ denote definiteness relationships between Hermitian matrices.
\end{enumerate}

\section{Background} 
\begin{figure}[t!]
\vspace{1.4ex}
\begin{center}
\includegraphics{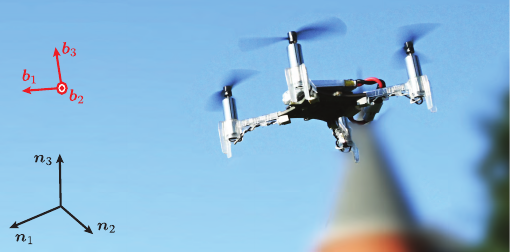}
\end{center}
\vspace{-2ex}
\caption{\hspace{-2ex}\textbf{Photograph of the quadrotor---a \mbox{Crazyflie\,$\bs{2.1}$}---used \mbox{in~the} flight tests performed to assess and demonstrate the \mbox{functionality} and performance of the proposed attitude control approach.} Here, \mbox{$\bs{\mathcal{N}} = \left\{\bs{n}_1, \bs{n}_2, \bs{n}_3\right\}$} and \mbox{$\bs{\mathcal{B}} = \left\{\bs{b}_1, \bs{b}_2, \bs{b}_3\right\}$} denote the inertial and \mbox{body-fixed} frames used for kinematic description and dynamic modeling. As customary, $\bs{\mathcal{N}}$ is fixed to the planet Earth and the origin of $\bs{\mathcal{B}}$ coincides with the CoM of the controlled rotational system. \label{Fig01}}
\vspace{-2ex}
\end{figure}

\vspace{-1ex}
\label{Section02}
In this section, we briefly review the attitude dynamics of a generic \mbox{rigid-body} rotational system and the basic structure of the \mbox{quaternion-based} controller used as the starting point for the presented research and as a benchmark. As indicated in \mbox{Fig.\,\ref{Fig01}}, \mbox{$\bs{\mathcal{N}} = \left\{\bs{n}_1, \bs{n}_2, \bs{n}_3\right\}$} and \mbox{$\bs{\mathcal{B}} = \left\{\bs{b}_1, \bs{b}_2, \bs{b}_3\right\}$} denote the inertial and \mbox{body-fixed} frames of reference used for kinematic and dynamic modeling, respectively. As is customary, $\bs{\mathcal{N}}$ is fixed to the planet Earth, and the origin of $\bs{\mathcal{B}}$ coincides with the \textit{center of mass} (CoM) of the modeled rotational system. As discussed in\mbox{\cite{BenaRM2022,BenaRM2023I}}, using quaternions and Euler's second law, we can describe the \mbox{open-loop} rotational dynamics of the modeled system as
\begin{align}
\begin{split}
\bs{\dot{\qbar}} &= \frac{1}{2} \bs{\qbar}\otimes
\begin{bmatrix}
0 \\
\bs{\omega} 
\end{bmatrix},\\
\bs{\dot{\omega}} &=  \bs{J}^{-1}\left[\bs{\tau}-\bs{\omega}\times \bs{J}\bs{\omega}\right],
\label{EQ01}
\end{split}
\end{align}
where $\bs{\qbar}$ is a unit quaternion that represents the orientation of $\bs{\mathcal{B}}$ relative to $\bs{\mathcal{N}}$; $\bs{\omega}$ is the angular velocity of $\bs{\mathcal{B}}$ relative to $\bs{\mathcal{N}}$, written in $\bs{\mathcal{B}}$ coordinates; $\bs{J}$ is the inertia matrix of the modeled rotational system, written in $\bs{\mathcal{B}}$ coordinates; and, $\bs{\tau}$ is the torque applied to the system and, in closed loop, generated by a control law. As is well known, \mbox{$\bs{\qbar} = \left[m\,\,\bs{n}^T\right]^T$}, with \mbox{$m = \cos{\frac{\Theta}{2}}$} and \mbox{$\bs{n} = \bs{u}\sin{\frac{\Theta}{2}}$}, where $\bs{u}$ is the Euler rotation axis and $\Theta$ is the amount that $\bs{\mathcal{N}}$ must rotate about $\bs{u}$ to be aligned with $\bs{\mathcal{B}}$.

For the purposes of controller synthesis, we define the instantaneous AEQ of the modeled rotational system as
\begin{align}
\bs{\qbar}_\text{e} = \bs{\qbar}^{-1} \otimes \bs{\qbar}_\text{d},
\label{EQ02}
\end{align}
where the desired attitude quaternion, $\bs{\qbar}_{\ts d}$, represents the rotation from the inertial frame, $\bs{\mathcal N}$, to the desired \mbox{body-fixed} frame, $\bs{\mathcal B}_{\ts d}$. Consistently, \mbox{$\bs{\qbar}_{\ts e} = \left[m_{\ts e}~\bs{n}_{\ts e}^T\right]^T$} represents the rotation from the current \mbox{body-fixed} frame, $\bs{\mathcal{B}}$, to the desired \mbox{body-fixed} frame, $\bs{\mathcal B}_{\ts d}$. Namely, \mbox{$m_{\ts{e}} = \cos \frac{{\Theta}_{\ts{e}}}{2}$} and \mbox{$\bs{n}_{\ts{e}} = \bs{u}_{\ts{e}}\sin \frac{{\Theta}_{\ts{e}}}{2}$}, where $\bs{u}_{\ts{e}}$ is the Euler axis about which the controlled rotational system must be rotated an amount \mbox{${\Theta}_{\ts{e}} \in [0, 2\pi)$\,rad} to align $\bs{\mathcal{B}}$ with $\bs{\mathcal{B}}_\ts{d}$. In agreement with the requirements for controller synthesis imposed by the second line of (\ref{EQ01}), directly from (\ref{EQ02}), it follows that $\bs{u}_{\ts{e}}$ is written in $\bs{\mathcal{B}}$ coordinates. Also, for \mbox{real-time} implementation, we must compute the desired angular velocity of the controlled rotational system, written in $\bs{\mathcal{B}}$ coordinates. This computation can be performed in two steps. First, we compute the desired angular velocity of the system in $\bs{\mathcal{B}}_{\text{d}}$ coordinates as 
\begin{align}
\left[
\begin{array}{c}
0 \\
\vspace{-2ex}
\\
\hat{\bs{\omega}}_\text{d}
\end{array}
\right]
= 2\bs{\qbar}_\ts{d}^{-1} \otimes \bs{\dot{\qbar}}_\text{d}.
\label{EQ03}
\end{align}
Then, to obtain the desired angular velocity of the system in $\bs{\mathcal{B}}$ coordinates, we consecutively apply two transformations, $\bs{S}_\ts{d}$ and $\bs{S}^T$, as 
\begin{align}
\bs{\omega}_\text{d} = \bs{S}^T \bs{S}_\ts{d} \hat{\bs{\omega}}_\text{d},
\label{EQ04}
\end{align}
where $\bs{S}_\ts{d}$ transforms vectors from $\bs{\mathcal{B}}_\ts{d}$ to $\bs{\mathcal{N}}$ coordinates, and $\bs{S}^T$ transforms vectors from $\bs{\mathcal{N}}$ to $\bs{\mathcal{B}}$ coordinates.

Next, as a starting point for the new research presented in this paper, employing the definitions of $\bs{\qbar}_\text{e}$ and $\bs{\omega}_\text{d}$, we define the control torque input
\begin{align}
\bs{\tau}_{\ts{b}} = \bs{J}\left[k_{\Theta} \bs{n}_{\ts{e}} + k_{\bs{\omega}} \bs{\omega}_{\ts{e}} + \bs{\dot{\omega}}_{\ts{d}}\right] + \bs{\omega}\times\bs{J}\bs{\omega},
\label{EQ05}
\end{align}
where \mbox{$k_{\Theta},k_{\bs{\omega}} \in \mathbb{R}_{>0}$}; \mbox{$\bs{\omega}_{\ts{e}} = \bs{\omega}_{\ts{d}} - \bs{\omega}$} is the \mbox{angular-velocity} tracking error; $\bs{J}\bs{\dot{\omega}}_{\ts{d}}$ is a feedforward term that cancels the \mbox{left-hand} side of the second equation in \eqref{EQ01} and provides faster tracking performance; and, \mbox{$\bs{\omega}\times\bs{J}\bs{\omega}$} is a \mbox{feedback-linearization} term that cancels the nonlinearity in the second equation of (\ref{EQ01}). As shown in\cite{BenaRM2022}, the CL attitude dynamics resulting from substituting the \textit{\mbox{right-hand} side} (RHS) of (\ref{EQ05}) into (\ref{EQ01}) exhibit two \mbox{equilibria}, specified by the pairs \mbox{$\left\{ \bs{\qbar}_\ts{e}^{\ast},\bs{\omega}_\ts{e}^{\ast}\right\}$} and \mbox{$\{ \bs{\qbar}_\ts{e}^{\dagger},\bs{\omega}_\ts{e}^{\ast}\}$}, with \mbox{$\bs{\qbar}_\ts{e}^{\ast} = \left[+1~0~0~0\right]^T$}, \mbox{$\bs{\qbar}_\ts{e}^{\dagger} = \left[-1~0~0~0\right]^T$}, and \mbox{$\bs{\omega}_\ts{e}^{\ast} = \left[0~0~0\right]^T$}. These two points define the same kinematic situation, but with different stability properties---one asymptotically stable and the other unstable. As expected from basic nonlinear theory, given that at both fixed points it occurs that \mbox{$\bs{n}_{\ts{e}} = \bs{0}$}, if the system state were to exactly reach the unstable equilibrium, the control torque specified by (\ref{EQ05}) cannot compel the controlled rotational system to execute a \mbox{$2\pi$-rad} rotation and thus converge to the stable equilibrium point, which prevents global asymptotic stability in $\mathcal{SO}(3)$ from being enforced. In fact, it can be shown that it is impossible to ensure global asymptotic stability with a continuous control law in this case\cite{bhat2000topological,liberzon2003switching}. In practice, however, the CL dynamics are expected to behave as a globally asymptotically stable system because any disturbance, large or small, would move the state away from the unstable equilibrium.   

Another relevant observation regarding the CL system resulting from using the law specified by (\ref{EQ05}) is that the effect of the proportional term, $\bs{J}k_{\Theta}\bs{n}_{\ts{e}}$, greatly diminishes as the instantaneous CL AEQ moves away from the CL equilibrium AEQ, for rotational errors larger than \mbox{$\pi$\,rad}, because of the $\sin \frac{\Theta_{\ts{e}}}{2}$ factor in \mbox{$\bs{n}_{\ts{e}} = \bs{u}_{\ts{e}}\sin \frac{{\Theta}_{\ts{e}}}{2}$}. From simple inspection, it is clear that the proportional control effort is close to minimal when the system state is close to the unstable CL equilibrium AEQ (corresponding to a rotational error about the Euler axis close to \mbox{$2\pi$\,rad}), is maximum halfway between the stable and unstable equilibrium AEQs (corresponding to a rotational error about the Euler axis of \mbox{$\pi$\,rad}), and is minimal at the stable equilibrium point (corresponding to a rotational error about the Euler axis close to \mbox{$0$\,rad}). These observations regarding the dynamic behavior induced by the law specified by (\ref{EQ05}) prompted us to introduce a new attitude control approach, which we present in Section\,\ref{Section03}.

\section{Two New Axis--Angle Attitude Control Laws}
\vspace{-0.5ex}
\label{Section03}
\subsection{Controller Synthesis and Implementation}
\begin{figure}[t!]
\vspace{1.4ex}
\begin{center}
\includegraphics{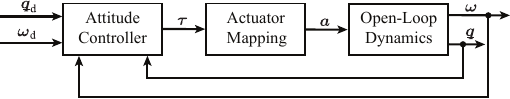}
\end{center}
\vspace{-2ex}
\caption{\hspace{-1ex}\textbf{Block diagram of the \mbox{upper-level} attitude control scheme con-sidered in this paper.} In this scheme, the attitude controller receives as inputs the desired attitude quaternion, $\bs{\qbarc}_{\ts{d}}$; the desired angular velocity, $\bs{\omega}_{\ts{d}}$; the measured attitude quaternion, $\bs{\qbarc}$; and, the measured angular velocity, $\bs{\omega}$. Using these inputs, at any given instant, it computes the control torque, \mbox{$\bs{\tau}\in\{\bs{\tau}_{\ts{b}}, \bs{\tau}_{1}, \bs{\tau}_{2}\}$}. The actuator mapping receives as its input $\bs{\tau}$ and generates as its output the vector signal $\bs{a}$ that excites the actuators of the \mbox{open-loop} dynamics of the controlled rotational system. \label{Fig02}}
\vspace{-2ex}
\end{figure}

\vspace{-0.5ex}
\label{Section03A}
A basic heuristic guideline for \mbox{feedback-controller} synthesis is to formulate a control law that increases the actuation effort as the control error grows\cite{KhalilHK2002,OgataK2010}, while ensuring the stability and robust performance of the resulting CL dynamics. Following this design philosophy to address the issues discussed in Section\,\ref{Section02} regarding \mbox{quaternion-based} attitude controllers, we propose a new type of attitude control law---to be integrated into the scheme shown in \mbox{Fig.\,\ref{Fig02}} for \mbox{real-time} implementation---with the form 
\begin{align}
\bs{\tau}_{j} = \bs{J}\left[k_{\Theta} \bs{p}_{\ts{e},j} + k_{\bs{\omega}} \bs{\omega}_{\ts{e}} + \bs{\dot{\omega}}_{\ts{d}}\right] + \bs{\omega}\times\bs{J}\bs{\omega},
\label{EQ06}
\end{align}
where the vector $\bs{p}_{\ts{e},j}$, for $j \in \mathbb{Z}_{>0}$, is defined by scaling the Euler axis associated with $\bs{\qbar}_\text{e}$ by a class $\mathcal{K}$ function of $\Theta_{\ts{e}}$ over the operational range $\left[0,2\pi\right)$. In this paper, we consider two functions,
\begin{align}
\bs{p}_{\ts{e},1} = \bs{u}_{\ts{e}} \frac{\Theta_{\ts{e}}}{2}~~~~\ts{and}~~~~\bs{p}_{\ts{e},2} = 2 \bs{u}_{\ts{e}} \sin \frac{\Theta_{\ts{e}}}{4}.
\label{EQ07}
\end{align}
Similarly to $\bs{n}_{\ts{e}}$, the generic \textit{scaled Euler axis} (SEA), $\bs{p}_{\ts{e},j}$, represents the direction about which the controlled rotational system must rotate to align $\bs{\mathcal{B}}$ with $\bs{\mathcal{B}}_{\ts{d}}$. The rationale behind selecting $\bs{p}_{\ts{e},1}$ is that the term scaling $\bs{u}_{\ts{e}}$ is directly proportional to the \mbox{Euler-axis} rotational error, $\Theta_{\ts{e}}$, and therefore the proportional control effort strictly increases with $\Theta_{\ts{e}}$. This approach, however, can lead to actuator saturation; for example, the quadrotor shown in \mbox{Fig.\,\ref{Fig01}} exhibits a measured \mbox{thrust-to-weight} ratio of only $2.4$, which implies that \mbox{high-gain} control laws almost certainly induce saturation of the rotor motors. 

Avoiding saturation is the main motivation for selecting the form of $\bs{p}_{\ts{e},2}$, which generates an increasing control effort as the \mbox{Euler-axis} rotational error, \mbox{$\Theta_{\ts{e}}$}, increases within the operational range \mbox{$\left[0,2\pi\right)$}, while reducing the likelihood of the system's actuators exceeding their operational limits. To see these advantages of $\bs{p}_{\ts{e},2}$ with respect to $\bs{p}_{\ts{e},1}$, note that the scaling factor in $\bs{\tau}_2$ is nonlinear in $\Theta_{\ts{e}}$ with a decreasing slope within the operational range \mbox{$\left[0,2\pi\right)$}. Accordingly, both laws, $\bs{\tau}_1$ and $\bs{\tau}_2$, generate similar control torques at small values of $\Theta_{\ts{e}}$, but the law that specifies $\bs{\tau}_2$ produces smaller control torques for relatively large values of $\Theta_{\ts{e}}$. The selection of $\sin\frac{\Theta_\ts{e}}{4}$ in the definition of $\bs{p}_{\ts{e},2}$ was inspired by attitude control laws based on the \textit{modified Rodrigues parameters}~(MRPs)\cite{SchaubH2018}---in which the SEA vector is defined as \mbox{$\bs{u}_{\ts{e}} \tan \frac{\Theta_{\ts{e}}}{4}$}. Note, however, that even though \mbox{MRPs-based} laws also generate an increasing control effort as the system state moves away from the stable CL equilibrium, the value of $\tan\frac{\Theta_{\ts{e}}}{4}$ grows unbounded as $\Theta_\ts{\ts{e}}$ approaches \mbox{$2\pi$\,rad}, which could be highly problematic in \mbox{real-time} applications. In contrast, $\bs{p}_{\ts{e},2}$ is well defined for any value of $\Theta_{\ts{e}}$. The inclusion of the scalars $\frac{1}{2}$ and $2$ in the respective formulations of $\bs{p}_{\ts{e},1}$ and $\bs{p}_{\ts{e},2}$ makes all three considered control laws---$\bs{\tau}_{\ts{b}}$, $\bs{\tau}_{1}$, and $\bs{\tau}_{2}$---comparable when the system operates near the stable CL equilibrium AEQ. 

\subsection{Equilibria of the Two Closed-Loop Systems}
\vspace{-0.5ex}
\label{Section03B}
Directly from substituting (\ref{EQ06}) and the left side of (\ref{EQ07}) into (\ref{EQ01}), it can be determined that the \mbox{state-space} representation of the CL rotational dynamics corresponding to the control torque $\bs{\tau}_1$ is given by 
\begin{align}
\begin{split}
\bs{\dot{\qbar}}_{\ts{e}} &= \frac{1}{2}
\left[
\begin{array}{c}
0 \\
\bs{\omega}_{\ts{e}}
\end{array}
\right]
\otimes \bs{\qbar}_{\ts{e}},\\
\bs{\dot{\omega}}_{\ts{e}} 
&= -\left[ k_{\Theta} \bs{u}_{\ts{e}} \frac{\Theta_{\ts{e}}}{2} + k_{\bs{\omega}} \bs{\omega}_{\ts{e}}\right], \\
\end{split}
\label{EQ08}
\end{align}
where $\frac{\Theta_{\ts{e}}}{2}$ is class $\mathcal{K}$ on \mbox{$\left[0, \infty \right)$} and, as a design restriction, \mbox{${\Theta}_{\ts{e}} \in \left[0,2\pi\right)$\,rad}. As explained in Section\,\ref{Section03C}, it can be shown that for \mbox{$k_{\Theta},k_{\bs{\omega}} \in \mathbb{R}_{>0}$}, the unique fixed point of this system is uniformly asymptotically stable. Consequently, an interpretation of this \mbox{state-space} representation is that it describes the dynamics of the deviation from the attitude and \mbox{angular-velocity} references, $\bs{\qbar}_{\ts{d}}$ and $\bs{\omega}_{\ts{d}}$, and by enforcing asymptotic stability of the CL fixed point, we ensure that \mbox{$\lim_{t \rightarrow \infty} \bs{\qbar} = \bs{\qbar}_{\ts{d}}$} and \mbox{$\lim_{t \rightarrow \infty} \bs{\omega} = \bs{\omega}_{\ts{d}}$}. An important additional characteristic of the system specified by (\ref{EQ08}) is that its RHS is discontinuous and, therefore, not Lipschitz continuous, which implies that a continuously differentiable solution does not exist\cite{KhalilHK2002}. Discontinuities occur at \mbox{$ \Theta_{\ts{e}} \in \left\{2\pi,4\pi,6\pi,\cdots\right\} = 2 \pi \ell$}, for \mbox{$\ell \in \mathbb{Z}_{>0}$}, as $\bs{u}_{\ts{e}}$ instantaneously switches sign at those values of $\Theta_{\ts{e}}$. However, similarly to the case presented in\cite{MayhewCG2011II}, using \mbox{Carath\'{e}odory's} concept of solution presented in\cite{HajekO1979}, it can be shown that an absolutely continuous function \mbox{$\bs{s}: I \rightarrow \mathcal{S}^3 \times \mathbb{R}^3$} satisfying (\ref{EQ08}) for almost every time \mbox{$t\in I \subset \mathbb{R}_{\geq 0}$} exists.

To find the fixed point(s) of the system specified by (\ref{EQ08}), we set \mbox{$\bs{\dot{\qbar}}_{\ts{e}} = [0\,\,0\,\,0\,\,0]^T$} and \mbox{$\bs{\dot{\omega}}_{\ts{e}} = [0\,\,0\,\,0]^T$}, and solve for $\bs{\qbar}_{\ts{e}}$ and $\bs{\omega}_{\ts{e}}$. Accordingly, we start with
\begin{align}
 \begin{split}
    -\frac{1}{2}\bs{n}_{\ts{e}}^T\bs{\omega}_{\ts{e}}  &= 0,\\
    - \frac{1}{2}\left[ \bs{n}_{\ts{e}} \times \bs{\omega}_{\ts{e}}  - m_{\ts{e}}\bs{\omega}_{\ts{e}}   \right] &= \bs{0}_{3\times1},\\
   -\left[k_{\Theta} \bs{u}_{\ts{e}} \frac{\Theta_{\ts{e}}}{2}+k_{\bs{\omega}} \bs{\omega}_{\ts{e}} \right] 
   &= \bs{0}_{3\times1},
\end{split}
\label{EQ09}   
\end{align}
and first notice that for the second equation to be satisfied, both vectors \mbox{$\bs{n}_{\ts{e}} \times \bs{\omega}_{\ts{e}}$} and \mbox{$m_{\ts{e}}\bs{\omega}_{\ts{e}}$} must be zero because they are orthogonal, which follows directly from the definition of cross product. For the first term in the second line to be zero, one of the following three propositions must hold:~(i)\,$\bs{n}_{\ts{e}}$ and $\bs{\omega}_{\ts{e}}$ are parallel;~(ii)\,$\bs{n}_{\ts{e}}$ and $\bs{\omega}_{\ts{e}}$ are both zero;~(iii)\,either $\bs{n}_{\ts{e}}$ or $\bs{\omega}_{\ts{e}}$ is zero. For the first equation to be satisfied, either (ii) or (iii) is true, or $\bs{n}_{\ts{e}}$ and $\bs{\omega}_{\ts{e}}$ are orthogonal. Since $\bs{n}_{\ts{e}}$ and $\bs{\omega}_{\ts{e}}$ cannot be simultaneously orthogonal and parallel, the only feasible option is either (ii) or (iii). Last, from simple examination of the third line, assuming $k_{\Theta},k_{\bs{\omega}}\in \mathbb{R}_{>0}$, it can be determined that if \mbox{$\bs{\omega}_{\ts{e}} = \bs{0}$}, the only possibility to satisfy this last relationship is to have \mbox{$\Theta_{\ts{e}} = 0$}, because $\bs{u}_{\ts{e}}$ is a unit vector (\mbox{$\bs{u}_{\ts{e}} \neq \bs{0}$}). Consistent with this argument, if $\Theta_{\ts{e}}=0$, the only possibility is to have \mbox{$\bs{\omega}_{\ts{e}}=\bs{0}$}. Therefore, noticing that $\Theta_{\ts{e}}=0$ implies that \mbox{$\bs{\qbar}_{\ts{e}}=[+1\,\,0\,\,0\,\,0]^T$}, we conclude that the unique solution to (\ref{EQ09}) and, therefore, the only fixed point of the CL system specified by (\ref{EQ08})---using AEQ representation---is the pair \mbox{$\left\{\bs{\qbar}^{\ast}_{\ts{e}}=[+1\,\,0\,\,0\,\,0]^T,\bs{\omega}^{\ast}_{\ts{e}}=[0\,\,0\,\,0]^T\right\}$}. As mentioned in \mbox{Section\,\ref{Section02}} and discussed in\cite{ChaturvediNA2011}, if a \mbox{time-invariant} continuous CL vector field on $\mathcal{SO}(3)$ has at least one isolated equilibrium, then it cannot be the only one; there must be at least two in order to satisfy \mbox{Poincar\'e--Hopf} \mbox{theorem\cite{KalabicU2016}}. However, discontinuous feedback laws can be used in attitude control to enforce the existence of a unique CL fixed point and achieve global stabilization on $\mathcal{SO}(3)$. Consistently, the discontinuity of the CL dynamics specified by (\ref{EQ08}) makes it is possible to have a unique fixed point in similar fashion to the cases discussed in\cite{MayhewCG2011II} and references therein.

Once more, directly from substituting (\ref{EQ06}) and the RHS of (\ref{EQ07}) into (\ref{EQ01}), it can be determined that the \mbox{state-space} representation of the CL rotational dynamics corresponding to the control torque $\bs{\tau}_2$, as specified by (\ref{EQ06}), is given by 
\begin{align}
\begin{split}
\bs{\dot{\qbar}}_{\ts{e}} &= \frac{1}{2}
\left[
\begin{array}{c}
0 \\
\bs{\omega}_{\ts{e}}
\end{array}
\right]
\otimes \bs{\qbar}_{\ts{e}},\\
\bs{\dot{\omega}}_{\ts{e}} 
&= -\left[ 2k_{\Theta} \bs{u}_{\ts{e}} \sin{\frac{\Theta_{\ts{e}}}{4}} + k_{\bs{\omega}} \bs{\omega}_{\ts{e}} \right], \\
\end{split}
\label{EQ10}
\end{align}
in which $\sin{\frac{\Theta_{\ts{e}}}{4}} $ is class $\mathcal{K}$ on $\left[0, 2\pi\right)$ and, as a design restriction, \mbox{${\Theta}_{\ts{e}} \in \left[0, 2\pi\right)$\,rad}. This design constraint is imposed to make coincide the range of operation with that over which $\sin{\frac{\Theta_{\ts{e}}}{4}}$ is class $\mathcal{K}$. As in the case specified by (\ref{EQ08}), to find the equilibrium point(s) of the \mbox{state-space} representation specified by (\ref{EQ10}), we set \mbox{$\bs{\dot{\qbar}}_{\ts{e}} = [0\,\,0\,\,0\,\,0]^T$} and \mbox{$\bs{\dot{\omega}}_{\ts{e}} = [0\,\,0\,\,0]^T$} and solve for $\bs{\qbar}_{\ts{e}}$ and $\bs{\omega}_{\ts{e}}$. Accordingly, we start with 
\begin{align}
 \begin{split}
    -\frac{1}{2}\bs{n}_{\ts{e}}^T\bs{\omega}_{\ts{e}}  &= 0,\\
    - \frac{1}{2}\left[ \bs{n}_{\ts{e}} \times \bs{\omega}_{\ts{e}}  - m_{\ts{e}}\bs{\omega}_{\ts{e}} \right] &= \bs{0}_{3\times1},\\
   -\left[ 2k_{\Theta} \bs{u}_{\ts{e}} \sin{\frac{\Theta_{\ts{e}}}{4}}+k_{\bs{\omega}} \bs{\omega}_{\ts{e}} \right] &= \bs{0}_{3\times1},
\end{split}
\label{EQ11}   
\end{align}
and first notice that the analysis for the first two equations is identical to that performed for (\ref{EQ08}) and (\ref{EQ09}). From simple examination of the third line, assuming \mbox{$k_{\Theta},k_{\bs{\omega}} \in \mathbb{R}_{>0}$}, it can be determined that \mbox{$\bs{\omega}_{\ts{e}} = \bs{0}$} \textit{if and only if} \mbox{$\Theta_{\ts{e}} = 4\pi \ell$}, with \mbox{$\ell \in \mathbb{Z}$}, because $\bs{u}_{\ts{e}}$ is a unit vector ($\bs{u}_{\ts{e}} \neq \bs{0}$). Therefore, over $\left[0,\infty \right)$ and the restricted range $\left[0,2\pi \right)\,\ts{rad}$, we conclude that the only equilibrium point of the CL system specified by (\ref{EQ10}) is the pair \mbox{$\left\{\bs{\qbar}^{\ast}_{\ts{e}}=[+1\,\,0\,\,0\,\,0]^T,\bs{\omega}^{\ast}_{\ts{e}}=[0\,\,0\,\,0]^T\right\}$}. This finding does not contradict the \mbox{Poincar\'e--Hopf} theorem---which states that a continuous time-invariant CL vector field on $\mathcal{SO}(3)$ cannot have only one isolated equilibrium\cite{KalabicU2016}---because the dynamics specified by (\ref{EQ10}) are discontinuous at \mbox{$ \Theta_{\ts{e}} \in \left\{2\pi,6\pi,10\pi,\cdots\right\} = 2 \pi (2\ell + 1) $}, for $\ell \in \mathbb{Z}_{\geq0}$. Furthermore, note that these discontinuities lie outside the selected range of operation, \mbox{$\left[0,2\pi \right)$}. 

\subsection{Stability Analysis}
\label{Section03C}
The stability of the equilibrium \mbox{$\left\{\bs{\qbar}^{\ast}_{\ts{e}}, \bs{\omega}^{\ast}_{\ts{e}}\right\}$} corresponding to the CL system specified by (\ref{EQ08}) can be analyzed by invoking \textit{Lyapunov's direct method} as stated in \mbox{Theorem\,4.9} of\cite{KhalilHK2002}. We state this result in the form of a proposition.
\begin{figure*}[t!]
\vspace{1.4ex}
\begin{center}
\includegraphics{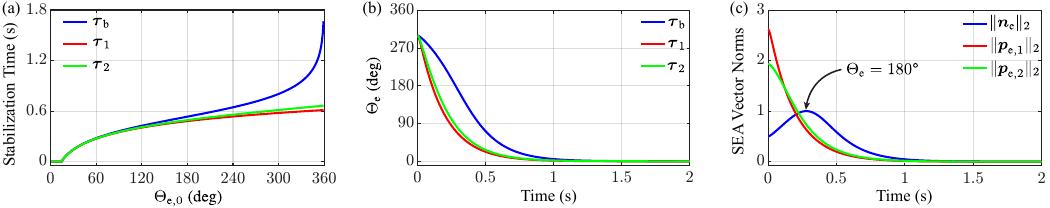}
\end{center}
\vspace{-3ex}
\caption{\textbf{Simulation results.}~\textbf{(a)}~Stabilization times versus the initial \mbox{Euler-axis} rotational error, $\Theta_{\ts{e},0}$, obtained using the three tested attitude control laws, $\bs{\tau}_{\ts{b}}$, $\bs{\tau}_1$, and $\bs{\tau}_2$.~\textbf{(b)}~Regulation responses of the \mbox{Euler-axis} rotational errors, $\Theta_{\ts{e}}(t)$, for \mbox{$t \in \left[0,2\right]$\,s} and initial condition \mbox{$\Theta_{\ts{e},0} = 300$\hspace{0.1ex}\textdegree}, obtained using the three tested control laws, $\bs{\tau}_{\ts{b}}$, $\bs{\tau}_1$, and $\bs{\tau}_2$.~\textbf{(c)}~SEA vector norms---$\|\bs{n}_{\ts{e}}\|_2$, $\|\bs{p}_{\ts{e},1}\|_2$, and $\|\bs{p}_{\ts{e},2}\|_2$---corresponding to the flight cases in~(b), evolving over time within the range \mbox{$\left[0,2\right]$\,s}. \label{Fig03}}
\vspace{-2ex}
\end{figure*}

\vspace{1ex}
\hspace{-2.2ex}\textbf{Proposition\,1.} \textit{Let the attitude and angular-velocity references}, $\bs{\qbar}_\ts{d}$ \textit{and} $\bs{\omega}_\ts{d}$, \textit{be smooth  and bounded functions of time, and let} $k_{\Theta}$ \textit{and} $k_{\bs{\omega}}$ \textit{be constant positive scalars}. \textit{Then, the fixed point} $\{\bs{\qbar}^{\ast}_{\ts{e}}, \bs{\omega}^{\ast}_{\ts{e}}\}$, \textit{with} \mbox{$\bs{\qbar}_{\ts{e}}^{\ast} = \left[+1\,\, 0\,\, 0\,\, 0 \right]^T$} \textit{and} \mbox{$\bs{\omega}_{\ts{e}}^{\ast} =  \left[ 0\,\, 0\,\, 0 \right]^T$,} \textit{of the CL \mbox{state-space} representation of the rotational dynamics specified by (\ref{EQ08}) is uniformly asymptotically stable.}

\vspace{1ex}
\hspace{-2.2ex}\textit{Proof.} We start by defining 
\begin{align}
V_1 = \frac{1}{2} k^{-1}_{\Theta}\bs{\omega}_{\ts{e}}^T\bs{\omega}_{\ts{e}} + \frac{1}{4}\Theta^2_{\ts{e}} + c \left(\bs{n}^T_{\ts{e}}\bs{n}_{\ts{e}} + \bs{n}^T_{\ts{e}}\bs{\omega}_{\ts{e}} \right) ,
\label{EQ12}
\end{align}
in which $c \in \mathbb{R}_{>0}$. It can be shown that $V_1$ meets the conditions specified in \cite{KhalilHK2002} for a \textit{Lyapunov function} (LF) because $\bs{\qbar}_\ts{d}$ and $\bs{\omega}_\ts{d}$ are smooth functions of time. Specifically, by imposing this property on the attitude and \mbox{angular-velocity} references, we ensure that $V_1$ is continuously differentiable in the selected range of operation, \mbox{$\Theta_{\ts{e}} \in \left[0,2\pi\right)$\,rad}, and when we substitute \mbox{$\{\bs{\qbar}^{\ast}_{\ts{e}}, \bs{\omega}^{\ast}_{\ts{e}}\}$} into (\ref{EQ12}), we obtain that \mbox{$V_1 \left(\bs{\qbar}^{\ast}_{\ts{e}}, \bs{\omega}^{\ast}_{\ts{e}} \right) = 0$}. Furthermore, $V_1$ can be lower bounded according to 
\begin{align}
\begin{split}
V_1 &\geq \frac{1}{2}k^{-1}_{\Theta}\|\bs{\omega}_{\ts{e}}\|^2_2 + c\|\bs{n}_{\ts{e}}\|_2^2-c\|\bs{n}_{\ts{e}}\|_2\|\bs{\omega}_{\ts{e}}\|_2\\
&= \bs{x}_{\ts{e}}^T\bs{P}\bs{x}_{\ts{e}}\geq \lambda_{\ts{min}}\left\{\bs{P}\right\}\|\bs{x}_{\ts{e}}\|_2^2,
\end{split}
\label{EQ13}
\end{align}
where 
\vspace{-2ex}
\begin{align}
\begin{split}
\bs{x}_{\ts{e}} &= \left[\|\bs{n}_{\ts{e}}\|_2~\|\bs{\omega}_{\ts{e}}\|_2\right]^T  \\
& \ts{and} \\
\bs{P} &= 
\left[
\begin{array}{cc}
\,c & -\frac{1}{2}c \\
\vspace{-2ex}
\\
-\frac{1}{2}c & ~~\,\frac{1}{2}k^{-1}_{\Theta}
\end{array} 
\right]
\succ \bs{0},
\end{split}
\label{EQ14}
\end{align}
for a value of $c$ sufficiently small, thus making $V_1$ strictly positive for any \mbox{$\bs{x}_{\ts{e}} \neq \bs{0}$}.

Next, taking the time derivative of $V_1$ in (\ref{EQ12}) yields
\begin{align}
\begin{split}
\largedot[1.4]{V}_1 &= -\bs{p}_{\ts{e},1}^T\bs{\omega}_{\ts{e}} - k_{\Theta}^{-1}k_{\bs{\omega}}\bs{\omega}_{\ts{e}}^T\bs{\omega}_{\ts{e}} + \frac{1}{2}\largedot[1.4]{\Theta}_{\ts{e}}\Theta_{\ts{e}} \\ &\hspace{3ex}+ 2c\bs{\dot{n}}^T_{\ts{e}}\bs{n}_{\ts{e}}+c\bs{\dot{n}}_{\ts{e}}^T\bs{\omega}_{\ts{e}}+c\bs{n}_{\ts{e}}^T\bs{\dot{\omega}}_{\ts{e}},
\end{split}
\label{EQ15}
\end{align}
which---following the analysis in Appendix\,A---can be upper bounded according to
\begin{align}
\begin{split}
\largedot[1.4]{V}_1 &\leq -k_{\Theta}^{-1}k_{\bs{\omega}}\|\bs{\omega}_{\ts{e}}\|_2^2 + c\|\bs{\omega}_{\ts{e}}\|_2\|\bs{n}_{\ts{e}}\|_2\\
&\hspace{3ex}+\frac{1}{2}c\|\bs{\omega}_{\ts{e}}\|_2^2 - ck_{\Theta}\|\bs{n}_{\ts{e}}\|_2^2 + ck_{\bs{\omega}}\|\bs{\omega}_{\ts{e}}\|_2\|\bs{n}_{\ts{e}}\|_2\\
&= -\bs{x}_{\ts{e}}^T\bs{Q}\bs{x}_{\ts{e}}\leq -\lambda_{\ts{min}}\left\{\bs{Q}\right\}\|\bs{x}_{\ts{e}}\|_2^2,
\end{split}
\label{EQ16}
\end{align}
with
\begin{align}
\bs{Q} = 
\left[
\begin{array}{cc}
ck_{\Theta} & -\frac{1}{2}c\left(k_{\bs{\omega}}+1\right) \\
\vspace{-2ex}
\\
-\frac{1}{2}c\left(k_{\bs{\omega}}+1\right) & ~~~k_{\Theta}^{-1}k_{\bs{\omega}}-\frac{1}{2}c
\end{array} 
\right]
\succ \bs{0},
\label{EQ17}
\end{align}
for a value of $c$ sufficiently small, thus making $\largedot[1.4]{V}_1$ strictly negative for any \mbox{$\bs{x}_{\ts{e}} \neq \bs{0}$}. 

Hence, we conclude that the unique CL equilibrium of the \mbox{state-space} representation specified by (\ref{EQ08}), \mbox{$\{\bs{\qbar}^{\ast}_{\ts{e}}, \bs{\omega}^{\ast}_{\ts{e}}\}$} is uniformly asymptotically stable.\hfill $\square$

\vspace{1ex}
Next, we analyze and prove the stability of the equilibrium \mbox{$\left\{\bs{\qbar}^{\ast}_{\ts{e}}, \bs{\omega}^{\ast}_{\ts{e}}\right\}$} corresponding to the CL system specified by (\ref{EQ10}), also invoking \textit{Lyapunov's direct method} as stated in \mbox{Theorem\,4.9} of\cite{KhalilHK2002}. We state this result in the form of a proposition.

\vspace{1ex}
\hspace{-2.2ex}\textbf{Proposition\,2.} \textit{Let the attitude and \mbox{angular-velocity} references}, $\bs{\qbar}_\ts{d}$ \textit{and} $\bs{\omega}_\ts{d}$, \textit{be smooth and bounded functions of time, and let} $k_{\Theta}$ \textit{and} $k_{\bs{\omega}}$ \textit{be constant positive scalars}. \textit{Then, the fixed point} \mbox{$\{\bs{\qbar}^{\ast}_{\ts{e}}, \bs{\omega}^{\ast}_{\ts{e}}\}$}, \textit{with} \mbox{$\bs{\qbar}_{\ts{e}}^{\ast} = \left[+1\,\, 0\,\, 0\,\, 0 \right]^T$} \textit{and} \mbox{$\bs{\omega}_{\ts{e}}^{\ast} =  \left[ 0\,\, 0\,\, 0 \right]^T$}, \textit{of the CL \mbox{state-space} representation of the rotational dynamics given by (\ref{EQ10}) is uniformly asymptotically stable inside the range of operation,} \mbox{$\Theta_{\ts{e}} \in \left[0,2\pi\right)$\,rad}.

\vspace{1ex}
\hspace{-2.2ex}\textit{Proof.} We start by defining 
\begin{align}
\begin{split}
V_2 = \frac{1}{2} k^{-1}_{\Theta}\bs{\omega}_{\ts{e}}^T\bs{\omega}_{\ts{e}} + 16\sin^2 \frac{\Theta_{\ts{e}}}{8} + c\bs{n}^T_{\ts{e}}\bs{n}_{\ts{e}}+ c\bs{n}^T_{\ts{e}}\bs{\omega}_{\ts{e}},
\end{split}
\label{EQ18}
\end{align}
in which \mbox{$c \in \mathbb{R}_{>0}$}. It can be shown that $V_2$ meets the conditions specified in \cite{KhalilHK2002} for an LF because $\bs{\qbar}_\ts{d}$ and $\bs{\omega}_\ts{d}$ are smooth functions of time. Specifically, by imposing this property on the attitude and \mbox{angular-velocity} references, we ensure that $V_2$ is continuously differentiable, and when we substitute \mbox{$\{\bs{\qbar}^{\ast}_{\ts{e}}, \bs{\omega}^{\ast}_{\ts{e}}\}$} into (\ref{EQ18}), we obtain that \mbox{$V_2 \left(\bs{\qbar}^{\ast}_{\ts{e}}, \bs{\omega}^{\ast}_{\ts{e}} \right) = 0$}. Furthermore, $V_2$ can be lower bounded according to
\begin{align}
\begin{split}
V_2 &\geq \frac{1}{2}k^{-1}_{\Theta}\|\bs{\omega}_{\ts{e}}\|^2_2 + c\|\bs{n}_{\ts{e}}\|_2^2-c\|\bs{n}_{\ts{e}}\|_2\|\bs{\omega}_{\ts{e}}\|_2\\
&= \bs{x}_{\ts{e}}^T\bs{P}\bs{x}_{\ts{e}}\geq \lambda_{\ts{min}}\left\{\bs{P}\right\}\|\bs{x}_{\ts{e}}\|_2^2,
\end{split}
\label{EQ19}
\end{align}
for a value of $c$ sufficiently small to make $V_2$ strictly positive for any \mbox{$\bs{x}_{\ts{e}} \neq \bs{0}$}. In (\ref{EQ19}), both $\bs{x}_{\ts{e}}$ and $\bs{P}$ have exactly the same form specified in (\ref{EQ14}). Next, taking the time derivative of $V_2$ defined by (\ref{EQ18}) yields
\begin{align}
\begin{split}
\hspace{-4ex}\largedot[1.4]{V}_2 &= -\bs{p}_{\ts{e},2}^T\bs{\omega}_{\ts{e}} -k_{\Theta}^{-1}k_{\bs{\omega}}\bs{\omega}_{\ts{e}}^T\bs{\omega}_{\ts{e}} + 32 \frac{\largedot[1.4]{\Theta}_{\ts{e}}}{8} \sin\frac{\Theta_{\ts{e}}}{8} \cos \frac{\Theta_{\ts{e}}}{8}  \\ 
&\hspace{3ex} + 2c\bs{\dot{n}}^T_{\ts{e}}\bs{n}_{\ts{e}}+c\bs{\dot{n}}_{\ts{e}}^T\bs{\omega}_{\ts{e}}+c\bs{n}_{\ts{e}}^T\bs{\dot{\omega}}_{\ts{e}},
\end{split}
\label{EQ20}
\end{align}
which---following the analysis in Appendix\,B---can be upper bounded according to
\begin{align}
\begin{split}
\largedot[1.4]{V}_2 &\leq -k_{\Theta}^{-1}k_{\bs{\omega}}\|\bs{\omega}_{\ts{e}}\|_2^2 + c\|\bs{\omega}_{\ts{e}}\|_2\|\bs{n}_{\ts{e}}\|_2\\
&\hspace{3ex}+\frac{1}{2}c\|\bs{\omega}_{\ts{e}}\|_2^2 - ck_{\Theta}\|\bs{n}_{\ts{e}}\|_2^2 + ck_{\bs{\omega}}\|\bs{\omega}_{\ts{e}}\|_2\|\bs{n}_{\ts{e}}\|_2\\
&= -\bs{x}_{\ts{e}}^T\bs{Q}\bs{x}_{\ts{e}}\leq -\lambda_{\ts{min}}\left\{\bs{Q}\right\}\|\bs{x}_{\ts{e}}\|_2^2,
\end{split}
\label{EQ21}
\end{align}
with $\bs{Q}$ defined exactly as in (\ref{EQ17}), for a value of $c$ sufficiently small to make $\largedot[1.4]{V}_2$ strictly negative for any \mbox{$\bs{x}_{\ts{e}} \neq \bs{0}$}. 

Hence, we conclude that the unique CL equilibrium of the \mbox{state-space} representation specified by (\ref{EQ10}), \mbox{$\{\bs{\qbar}^{\ast}_{\ts{e}}, \bs{\omega}^{\ast}_{\ts{e}}\}$}, is uniformly asymptotically stable.\hfill $\square$

\section{Simulations and Experiments}
\vspace{-0.5ex}
\label{Section04}
\subsection{Numerical Simulations}
\vspace{-0.5ex}
\label{Subsection04A}
To first assess and demonstrate the functionality and performance of the two attitude control laws specified by (\ref{EQ06}) and (\ref{EQ07}), we compared them with the benchmark law specified by (\ref{EQ05}) through numerical simulations implemented and run on \mbox{Simulink\,$23.2$}~(\mbox{MATLAB\,R$2023$b}), using the \mbox{Dormand-Prince} algorithm with a fixed step size of $10^{-4}\,\ts{s}$. We set and initialized the simulations using the parameters of the \mbox{Crazyflie\,$2.1$}\cite{bitcraze}---shown in \mbox{Fig.\,\ref{Fig01}}---and \mbox{empirically-selected} controller gains that satisfy the stability conditions discussed in \mbox{Section\,\ref{Section03}}. Accordingly, we selected \mbox{$\bs{J} = \ts{diag}\{16.57,16.66,29.26\}\hspace{-0.2ex}\cdot\hspace{-0.2ex}10^{-6}\,\ts{kg}\hspace{-0.2ex}\cdot\hspace{-0.2ex}\ts{m}^2$}, \mbox{$k_{\Theta} = 1000~\ts{N} \cdot \ts{m}$}, and \mbox{$k_{\bs{\omega}} = 100~\ts{N} \cdot  \ts{m} \cdot \ts{s} \cdot \ts{rad}^{-1}$}. It can be verified that for these controller gains, a value \mbox{$c\leq10^{-3}$} makes $\bs{P}$ and $\bs{Q}$ positive definite. For each control law---$\bs{\tau}_{\ts{b}}$, $\bs{\tau}_{1}$, and $\bs{\tau}_{2}$---we simulated $359$ \mbox{tumble-recovery} maneuvers during which the flier is commanded to stabilize itself from an upended flight position to reach the desired state given by the pair \mbox{$\left\{\bs{\qbar}_{\ts{d}} = [+1\,\,0\,\,0\,\,0]^T,\bs{\omega}_{\ts{d}} = [0\,\,0\,\,0]^T\right\}$}. Specifically, each simulation was initialized at a time $t_0$ with an Euler axis of rotation, \mbox{$\bs{u}_0 = \bs{u}(t_0)$}, generated randomly with a new seed using MATLAB. The initial rotation about the Euler axis, \mbox{$\Theta_0 = \Theta(t_0)$}, was sequentially selected from $1$ to \mbox{$359$\hspace{0.1ex}\textdegree} for each new simulation in increments of \mbox{$1$\hspace{0.1ex}\textdegree}, and the initial angular velocity, \mbox{$\bs{\omega}_0 = \bs{\omega}(t_0)$}, was set to zero for all the simulated cases. To evaluate and compare the controllers, we use as a figure of merit the \textit{stabilization time}, defined as the time it takes for the rotation error, $\Theta_{\ts{e}}$, about the \mbox{attitude-error} Euler axis, $\bs{u}_{\ts{e}}$, to reach a value lower than \mbox{$15$\hspace{0.1ex}\textdegree}.

\mbox{Fig.\,\ref{Fig03}(a)} shows the stabilization times corresponding to the three simulated control laws as functions of the initial rotation error, \mbox{$\Theta_{\ts{e},0} = \Theta_{\ts{e}}(t_0)$}. As seen, the results corresponding to the three different controllers are very similar for \mbox{$\Theta_{\ts{e},0} \leq 180$\hspace{0.1ex}\textdegree}; however, the superior performance of laws $\bs{\tau}_1$ and $\bs{\tau}_2$, relative to that corresponding to the \mbox{quaternion-based} $\bs{\tau}_{\ts{b}}$, becomes evident for cases in which \mbox{$\Theta_{\ts{e},0} > 180$\hspace{0.1ex}\textdegree}. For example, for values of \mbox{$\Theta_{\ts{e},0} > 350$\hspace{0.1ex}\textdegree}, the stabilization time achieved with $\bs{\tau}_{\ts{b}}$ is more than $2$ times higher than those achieved with the two laws introduced and studied in this paper. This difference in performance can be critical in situations in which the controlled system must recover from an unexpected disturbance that significantly changed its attitude. Furthermore, regarding design, functionality, and performance, it is important to note that $\bs{\tau}_1$ and $\bs{\tau}_2$ were conceived to be implemented as constituents of switching control schemes such as those presented in\cite{GoncalvesFMFR2024I,GoncalvesFMFR2024II,GoncalvesFMFR2024III}. Consequently, the rationale behind testing cases in which \mbox{$\Theta_{\ts{e},0}>180$\hspace{0.1ex}\textdegree} is that, in some cases during flight, it is expected the switching controller to apply the proportional torque component in the direction of the longest \mbox{rotational-error} path.

\mbox{Figs.\,\ref{Fig03}(b)~and~(c)} show the simulated regulation responses of the rotation error, $\Theta_{\ts{e}}(t)$, for \mbox{$t \in \left[0,2\right]\,\ts{s}$}, about the \mbox{attitude-error} Euler axis, and the SEA vector norms---$\|\bs{n}_{\ts{e}}\|_2$, $\|\bs{p}_{\ts{e},1}\|_2$, and $\|\bs{p}_{\ts{e},2}\|_2$---in the proportional terms of the three compared laws---$\bs{\tau}_{\ts{b}}$, $\bs{\tau}_{1}$, and $\bs{\tau}_2$---for \mbox{$\Theta_{\ts{e},0}=300$\hspace{0.1ex}\textdegree}. Simple inspection of \mbox{Fig.\,\ref{Fig03}(b)} allows us to determine that the stabilization time in the \mbox{quaternion-based} control case is significantly higher than in the \mbox{$\bs{\tau}_1$-based} and \mbox{$\bs{\tau}_2$-based} cases; \mbox{$0.80$\,s} versus $0.58$ and \mbox{$0.61$\,s}, respectively. The data in \mbox{Fig.\,\ref{Fig03}(c)} confirms that the effect sought with the introduction of the two laws specified by (\ref{EQ06}) and (\ref{EQ07}) is fully accomplished as the proportional control effort applied to the rotational system at \mbox{$\Theta_{\ts{e}} = 300$\hspace{0.1ex}\textdegree} is about more than three and two times those corresponding to $\bs{\tau}_1$ and $\bs{\tau}_2$, respectively. Furthermore, both $\|\bs{p}_{\ts{e},1}\|_2$ and $\|\bs{p}_{\ts{e},2}\|_2$ decrease as \mbox{$\Theta_{\ts{e}}$} decreases from $300$ to \mbox{$0$\hspace{0.1ex}\textdegree}. In contrast, it can be observed that the magnitude of $\|\bs{n}_{\ts{e}}\|_2$ increases as \mbox{$\Theta_{\ts{e}}$} decreases from $300$ to \mbox{$180$\hspace{0.1ex}\textdegree} to rapidly decrease after passing this point. This dynamical behavior is inconsistent with the notion that \textit{the actuation effort should grow with the magnitude of the control error}.

The simulated maneuvers presented in this section were selected to demonstrate and highlight the differences between the two new control laws---specified by (\ref{EQ06}) and (\ref{EQ07})---and the \mbox{quaternion-based} law used as benchmark---specified by (\ref{EQ05}). By choosing an initial condition of the rotation error, $\Theta_{\ts{e},0}$, about the \mbox{attitude-error} Euler axis larger than \mbox{$180$\hspace{0.1ex}\textdegree}, the CL AEQ of the attitude dynamics resulting from using (\ref{EQ05}) lies very close to the unstable CL equilibrium AEQ and, therefore, as explained in \mbox{Section\,\ref{Section02}}, the proportional control effort takes values significantly lower than the maximum achievable. In contrast, by design, the two new control laws always generate a proportional driving torque that increases with $\Theta_{\ts{e}}$. A last interesting observation regarding the simulation results summarized in \mbox{Fig.\,\ref{Fig03}} is that the time evolutions of $\Theta_{\ts{e}}$ corresponding to $\bs{\tau}_1$ and $\bs{\tau}_2$ look very similar. The values of $\|\bs{p}_{\ts{e},1} \|_2$ and $\|\bs{p}_{\ts{e},2} \|_2$, however, look very different at large magnitudes of $\Theta_{\ts{e}}$. The relatively high values of $\|\bs{p}_{\ts{e},1} \|_2$ at high values of $\Theta_{\ts{e}}$ might lead to actuator saturation depending on the characteristics of the controlled rotational system. For this reason, for the \mbox{real-time} flight control experiments presented next, we implemented $\bs{\tau}_{\ts{b}}$ and $\bs{\tau}_2$ only, considering the limited \mbox{thrust-to-weight} of the tested quadrotor, i.e., $2.4$.
\begin{figure}[t!]
\vspace{1.4ex}
\begin{center}
\includegraphics{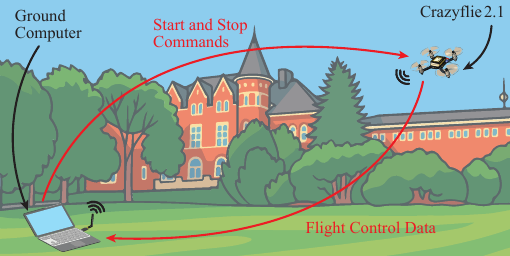}
\end{center}
\vspace{-2ex}
\caption{\textbf{Experimental setup used during the performance of flight control tests.} All the experiments were performed outdoor using a ground computer equipped with a \mbox{Crazyradio\,$2.0$} dongle, which sends the initialization and stop commands for the execution of controlled flight maneuvers. Also, during flight, the flier---a \mbox{Crazyflie\,$2.1$}---sends its instantaneous  state to the ground computer for data collection. \label{Fig04}}
\vspace{-2ex}
\end{figure}
\begin{figure*}[t!]
\vspace{1.4ex}
\begin{center}
\includegraphics{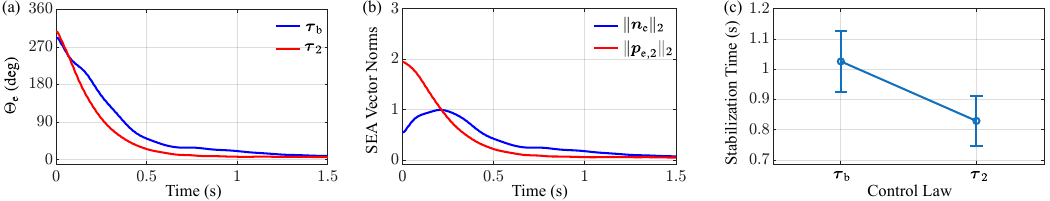}
\end{center}
\vspace{-3ex}
\caption{\textbf{Experimental results.}~\textbf{(a)}~Time evolutions of the \mbox{Euler-axis} rotational errors, $\Theta_{\ts{e}}(t)$, for \mbox{$t \in  \left[0,1.5\right]$\,s}, obtained using the benchmark attitude control law, $\bs{\tau}_{\ts{b}}$, and the new attitude control law, $\bs{\tau}_2$.~\textbf{(b)}~SEA vector norms---$\|\bs{n}_{\ts{e}}\|_2$, $\|\bs{p}_{\ts{e},1}\|_2$, and $\|\bs{p}_{\ts{e},2}\|_2$---corresponding to the flight cases in~(a), evolving over time within the range \mbox{$\left[0,1.5\right]$\,s}.~\textbf{(c)}~Mean and SEM of the stabilization times corresponding to $25$ \mbox{back-to-back} experiments respectively performed using $\bs{\tau}_{\ts{b}}$ and $\bs{\tau}_2$. \label{Fig05}}
\vspace{-2ex}
\end{figure*}

\subsection{Real-Time Flight Tests}
\vspace{-0.5ex}
\label{Subsection04B}
To compare the dynamical behavior of the two tested attitude control laws, $\bs{\tau}_{\ts{b}}$ and $\bs{\tau}_2$, in real time, we implemented each of them as a constituent of the \mbox{model-predictive} switching scheme described in\cite{GoncalvesFMFR2024I}, using the quadrotor shown in \mbox{Fig.\,\ref{Fig01}} with the same controller parameters specified in \mbox{Section\,\ref{Subsection04A}}. As depicted in Fig.\,\ref{Fig04}, we performed flight tests outdoor using a ground computer equipped with the \mbox{Crazyradio\,$2.0$} dongle. During an experiment, this ground system sends initialization and stop commands for executing maneuvers, and collects the flight data sent from the drone to the ground computer through radio communication at a rate of \mbox{$100$\,Hz}. The executed flight tests were conceived to emulate \mbox{high-speed} \mbox{tumble-recovery} maneuvers resulting from \textit{throw launches}---expected to be used in military, or search and rescue scenarios---consisting of two phases. During the first phase, the quadrotor is thrown by the experimenter in the air with an unknown relatively high angular velocity while the controller and rotors remain off. During the second phase, the controller is activated and the rotors are turned on in order to stabilize the quadrotor's attitude and reach a desired hovering state given by the pair \mbox{$\left\{\bs{\qbar}_{\ts{d}} = [+1\,\,0\,\,0\,\,0]^T,\bs{\omega}_{\ts{d}} = [0\,\,0\,\,0]^T\right\}$}. The quadrotor executes the entire maneuver autonomously, using onboard sensing and computation only, with the attitude controller running at a frequency of \mbox{$500$\,Hz}.

\mbox{Fig.\,\ref{Fig05}(a)} shows the time evolution of $\Theta_{\ts{e}}$ measured during two experiments correspondingly performed using $\bs{\tau}_{\ts{b}}$ and $\bs{\tau}_2$, and initialized with similar initial conditions. In these particular tests, similarly to the simulation cases presented in \mbox{Section\,\ref{Subsection04A}}, the stabilization time corresponding to $\bs{\tau}_2$ is significantly shorter than that achieved with $\bs{\tau}_{\ts{b}}$---$0.63$ versus \mbox{$1.18$\,s}. \mbox{Fig.\,\ref{Fig05}(b)} shows the time evolutions of $\|\bs{n}_{\ts{e}}\|_2$ and $\|\bs{p}_{\ts{e},2}\|_2$. Similarly to the simulation cases presented in \mbox{Section\,\ref{Subsection04A}}, $\|\bs{n}_{\ts{e}}\|_2$ increases as $\Theta_{\ts{e}}$ decreases from its initial condition until reaching \mbox{$180$\hspace{0.1ex}\textdegree}, and decreases after passing this point, thus exhibiting the undesired behavior already discussed in the previous sections. In contrast, $\|\bs{p}_{\ts{e},2}\|_2$ decreases as $\Theta_{\ts{e}}$ decreases from its initial condition until reaching a value of approximately \mbox{$0$\hspace{0.1ex}\textdegree}. Video footage of these experiments can be seen in the accompanying supplementary movie.

Last, in \mbox{Fig.\,\ref{Fig05}(c)}, each datum shows the mean and \textit{standard error of the mean} (SEM) of the stabilization times corresponding to $25$ \mbox{back-to-back} experiments respectively executed using $\bs{\tau}_{\ts{b}}$ (left) and $\bs{\tau}_2$ (right). As seen, the mean values achieved with $\bs{\tau}_{\ts{b}}$ and $\bs{\tau}_2$ are $1.03$ and $0.83$\,s, respectively. Although the difference between these two values is not pronounced due to actuation limitations of the experimental platform, they are consistent with the simulation data presented in \mbox{Section\,\ref{Subsection04A}} and provide convincing experimental evidence of the superior performance achievable with $\bs{\tau}_2$ relative to that obtained with $\bs{\tau}_{\ts{b}}$. Furthermore, the SEM values corresponding to $\bs{\tau}_{\ts{b}}$ and $\bs{\tau}_2$ are $0.10$ and $0.08$\,s, respectively, which suggests that the mean value corresponding to $\bs{\tau}_2$ is a more precise estimate of the stabilization time, used as a measured of performance. A large variability of the data is expected due to the \textit{stochastic} nature of \mbox{tumble-recovery} experiments; in some experiments, the initial $\Theta_{\ts{e}}$ and $\bs{\omega}_{\ts{e}}$ are small, which results in relatively low stabilization times, while in other experiments, the initial $\Theta_{\ts{e}}$ and $\bs{\omega}_{\ts{e}}$ are large, which results in relatively high stabilization times.

\section{Conclusions and Future Research}
\vspace{-0.5ex}
\label{Section05}
We introduced a new methodology for synthesizing attitude controllers for rotational systems that can be modeled as rigid bodies, such as satellites, drones, and microswimmers. In this approach---in contrast to \mbox{quaternion-based} schemes---the controller is defined using an Euler \mbox{axis--angle} \mbox{attitude-error} law that guarantees the existence of a unique CL equilibrium AEQ and provides greater flexibility in the use of \mbox{proportional-control} effort. To illustrate the proposed methodology, we defined two different attitude control laws and, for each case studied, we showed that the unique equilibrium of the CL dynamics is uniformly asymptotically stable by constructing a strict Lyapunov function for the system. To test and demonstrate the functionality and performance of the two resulting CL schemes, we conducted hundreds of numerical simulations and executed dozens of \mbox{real-time} flight experiments using a \mbox{small-sized} quadrotor (\mbox{Crazyflie\,$2.1$}). In particular, we studied and discussed \mbox{tumble-recovery} maneuvers controlled by the three controllers---the two new ones and a benchmark---presented in this paper. The obtained simulation and experimental results provide compelling evidence for the suitability of the new control schemes, which can be particularly useful as constituents of switching architectures that account for the angular velocity when selecting the direction in which proportional torque is applied. Currently, controllers of this type do not account for actuator saturation, which is a matter of ongoing and future research at the \textit{Autonomous\,Microrobotic\,Systems\,Laboratory}~(AMSL).

\bibliographystyle{IEEEtran}
\bibliography{paper}

{\small
\section*{\small Appendix}
\subsectionsmall{\small Upper Bound for $\largedot[1.4]{\,V_1}$ in \mbox{Proposition\,1}}
\noindent It is easy to verify that the time derivative of $V_1$ is
\begin{align}
\begin{split}
\largedot[1.4]{V}_1 &= -\bs{p}_{\ts{e},1}^T\bs{\omega}_{\ts{e}} - k_{\Theta}^{-1}k_{\bs{\omega}}\bs{\omega}_{\ts{e}}^T\bs{\omega}_{\ts{e}} + \frac{1}{2}\largedot[1.4]{\Theta}_{\ts{e}}\Theta_{\ts{e}} + 2c\bs{\dot{n}}^T_{\ts{e}}\bs{n}_{\ts{e}}
\\ &\hspace{3ex}+ c\bs{\dot{n}}_{\ts{e}}^T\bs{\omega}_{\ts{e}}+c\bs{n}_{\ts{e}}^T\bs{\dot{\omega}}_{\ts{e}}.
\end{split}
\label{EQ22}
\end{align}
Then, recalling that \mbox{$\bs{p}_{\ts{e},1} = \bs{u}_{\ts{e}}\frac{\Theta}{2}$} and \mbox{$\largedot[1.4]{\Theta}_{\ts{e}} = \bs{\omega}_{\ts{e}}^T\bs{u}_{\ts{e}}$}, we can rewrite (\ref{EQ22}) as
\begin{align}
\begin{split}
\largedot[1.4]{V}_1 &= -\frac{1}{2}\bs{u}_{\ts{e}}^T\bs{\omega}_{\ts{e}}\Theta_{\ts{e}} - k_{\Theta}^{-1}k_{\bs{\omega}}\bs{\omega}_{\ts{e}}^T\bs{\omega}_{\ts{e}} + \frac{1}{2}\bs{\omega}_{\ts{e}}^T\bs{u}_{\ts{e}}\Theta_{\ts{e}} \\   &\hspace{3ex}+2c\bs{\dot{n}}_{\ts{e}}^T\bs{n}_{\ts{e}}+c\bs{\dot{n}}_{\ts{e}}^T\bs{\omega}_{\ts{e}}+c\bs{n}_{\ts{e}}^T\bs{\dot{\omega}}_{\ts{e}}\\
&= -k_{\Theta}^{-1}k_{\bs{\omega}}\bs{\omega}_{\ts{e}}^T\bs{\omega}_{\ts{e}}+2c\bs{\dot{n}}_{\ts{e}}^T\bs{n}_{\ts{e}}+c\bs{\dot{n}}_{\ts{e}}^T\bs{\omega}_{\ts{e}}+c\bs{n}_{\ts{e}}^T\bs{\dot{\omega}}_{\ts{e}},
\end{split}
\label{EQ23}
\end{align}
in which, directly from (\ref{EQ08}),
\begin{align}
\begin{split}
\bs{\dot{n}}_{\ts{e}} = \frac{1}{2}\left(m_{\ts{e}}\bs{\omega_{\ts{e}}} + \bs{\omega}_{\ts{e}} \times \bs{n}_{\ts{e}}\right).
\end{split}
\label{EQ24}
\end{align}
Next, substituting (\ref{EQ24}) into (\ref{EQ23}) and noticing that \mbox{$\bs{\omega}_{\ts{e}} \times \bs{n}_{\ts{e}}$} is orthogonal to $\bs{\omega}_{\ts{e}}$ and $\bs{n}_{\ts{e}}$, (\ref{EQ23}) can be further simplified as 
\begin{align}
\begin{split}
\hspace{-2ex}
\largedot[1.4]{V}_1 &= -k_{\Theta}^{-1}k_{\bs{\omega}}\bs{\omega}_{\ts{e}}^T\bs{\omega}_{\ts{e}}+cm_{\ts{e}}\bs{\omega}_{\ts{e}}^T\bs{n}_{\ts{e}}+\frac{1}{2}cm_{\ts{e}}\bs{\omega}_{\ts{e}}^T\bs{\omega}_{\ts{e}}+c\bs{n}_{\ts{e}}^T\bs{\dot{\omega}}_{\ts{e}},
\end{split}
\label{EQ25}
\end{align}
which, by using the RHS of the second equation in (\ref{EQ08}), becomes 
\begin{align}
\begin{split}
\largedot[1.4]{V}_1 &= -k_{\Theta}^{-1}k_{\bs{\omega}}\bs{\omega}_{\ts{e}}^T\bs{\omega}_{\ts{e}}+cm_{\ts{e}}\bs{\omega}_{\ts{e}}^T\bs{n}_{\ts{e}}+\frac{1}{2}cm_{\ts{e}}\bs{\omega}_{\ts{e}}^T\bs{\omega}_{\ts{e}}\\
&\hspace{3ex}-ck_{\Theta}\bs{n}_{\ts{e}}^T\bs{p}_{\ts{e},1} -ck_{\bs{\omega}}\bs{n}_{\ts{e}}^T\bs{\omega}_{\ts{e}}.
\end{split}
\label{EQ26}
\end{align}
This expression can be upper bounded as
\begin{align}
\begin{split}
\largedot[1.4]{V}_1 &\leq -k_{\Theta}^{-1}k_{\bs{\omega}}\|\bs{\omega}_{\ts{e}}\|_2^2+c\|\bs{\omega}_{\ts{e}}\|_2\|\bs{n}_{\ts{e}}\|_2+\frac{1}{2}c\|\bs{\omega}_{\ts{e}}\|_2^2\\
&\hspace{3ex}-ck_{\Theta}\|\bs{n}_{\ts{e}}\|_2^2 +ck_{\bs{\omega}}\|\bs{n}_{\ts{e}}\|_2\|\bs{\omega}_{\ts{e}}\|_2,
\end{split}
\label{EQ27}
\end{align}
because \mbox{$-\|\bs{n}_\ts{e}\|_2^2\geq -\|\bs{n}_{\ts{e}}\|_2\|\bs{p}_{\ts{e},1}\|_2$}, for any \mbox{$\Theta_{\ts{e}}\in \left[0,2\pi\right)$}---the operational range selected by design. Last, using matrix manipulation rules, we obtain   
\begin{align}
\begin{split}
\largedot[1.4]{V}_1 \leq 
-\bs{x}^T_{\ts{e}}
\left[
\begin{array}{cc}
ck_{\Theta} & -\frac{1}{2}c\left(k_{\bs{\omega}}+1\right) \\
\vspace{-2ex}
\\
-\frac{1}{2}c\left(k_{\bs{\omega}}+1\right) & ~~~k_{\Theta}^{-1}k_{\bs{\omega}}-\frac{1}{2}c
\end{array} 
\right]
\bs{x}_{\ts{e}},
\end{split}
\label{EQ28}
\end{align}
where \mbox{$\bs{x}_{\ts{e}} = \left[\|\bs{n}_{\ts{e}}\|_2~\|\bs{\omega}_{\ts{e}}\|_2\right]^T$}.

\subsectionsmall{\small Upper Bound for $\largedot[1.4]{\,V_2}$ in \mbox{Proposition\,2}}
\noindent It is easy to verify that the time derivative of $V_2$ is
\begin{align}
\begin{split}
\hspace{-2ex}
\largedot[1.4]{V}_2 &= -\bs{p}_{\ts{e},2}^T\bs{\omega}_{\ts{e}} - k_{\Theta}^{-1}k_{\bs{\omega}}\bs{\omega}_{\ts{e}}^T\bs{\omega}_{\ts{e}} + 32 \frac{\largedot[1.4]{\Theta}_{\ts{e}}}{8}\sin \frac{\Theta_{\ts{e}}}{8} \cos \frac{\Theta_{\ts{e}}}{8} \\ &\hspace{3ex}+ 2c\bs{\dot{n}}^T_{\ts{e}}\bs{n}_{\ts{e}}+c\bs{\dot{n}}_{\ts{e}}^T\bs{\omega}_{\ts{e}}+c\bs{n}_{\ts{e}}^T\bs{\dot{\omega}}_{\ts{e}},
\end{split}
\label{EQ29}
\end{align}
Then, recalling that \mbox{$\bs{p}_{\ts{e},2} = 2\bs{u}_{\ts{e}}\sin\left(\frac{\Theta}{4}\right)$}, \mbox{$\largedot[1.4]{\Theta}_{\ts{e}} = \bs{\omega}_{\ts{e}}^T\bs{u}_{\ts{e}}$}, and $\sin \frac{\Theta_{\ts{e}}}{8} \cos \frac{\Theta_{\ts{e}}}{8} = \frac{1}{2}\sin \frac{\Theta_{\ts{e}}}{4}$, we can rewrite (\ref{EQ29}) as 
\begin{align}
\begin{split}
\largedot[1.4]{V}_2 &= -2\bs{u}_{\ts{e}}^T\bs{\omega}_{\ts{e}}\sin\frac{\Theta_{\ts{e}}}{4} - k_{\Theta}^{-1}k_{\bs{\omega}}\bs{\omega}_{\ts{e}}^T\bs{\omega}_{\ts{e}} +2\bs{\omega}_{\ts{e}}^T\bs{u}_{\ts{e}}\sin\frac{\Theta_{\ts{e}}}{4} \\
&\hspace{3ex}+2c\bs{\dot{n}}_{\ts{e}}^T\bs{n}_{\ts{e}}+c\bs{\dot{n}}_{\ts{e}}^T\bs{\omega}_{\ts{e}}+c\bs{n}_{\ts{e}}^T\bs{\dot{\omega}}_{\ts{e}}\\
&= k_{\Theta}^{-1}k_{\bs{\omega}}\bs{\omega}_{\ts{e}}^T\bs{\omega}_{\ts{e}}+2c\bs{\dot{n}}_{\ts{e}}^T\bs{n}_{\ts{e}}+c\bs{\dot{n}}_{\ts{e}}^T\bs{\omega}_{\ts{e}}+c\bs{n}_{\ts{e}}^T\bs{\dot{\omega}}_{\ts{e}},
\end{split}
\label{EQ30}
\end{align}
in which, directly from (\ref{EQ10}),
\begin{align}
\begin{split}
\bs{\dot{n}}_{\ts{e}} = \frac{1}{2}\left(m_{\ts{e}}\bs{\omega_{\ts{e}}} + \bs{\omega}_{\ts{e}} \times \bs{n}_{\ts{e}}\right).
\end{split}
\label{EQ31}
\end{align}
Next, substituting (\ref{EQ31}) into (\ref{EQ30}) and noticing that \mbox{$\bs{\omega}_{\ts{e}} \times \bs{n}_{\ts{e}}$} is orthogonal to $\bs{\omega}_{\ts{e}}$ and $\bs{n}_{\ts{e}}$, (\ref{EQ30}) can be further simplified as 
\begin{align}
\begin{split}
\hspace{-2ex}\largedot[1.4]{V}_2 &= -k_{\Theta}^{-1}k_{\bs{\omega}}\bs{\omega}_{\ts{e}}^T\bs{\omega}_{\ts{e}}+cm_{\ts{e}}\bs{\omega}_{\ts{e}}^T\bs{n}_{\ts{e}}+\frac{1}{2}cm_{\ts{e}}\bs{\omega}_{\ts{e}}^T\bs{\omega}_{\ts{e}}+c\bs{n}_{\ts{e}}^T\bs{\dot{\omega}}_{\ts{e}},
\end{split}
\label{EQ32}
\end{align}
which, by using the RHS of the second equation in (\ref{EQ10}), becomes 
\begin{align}
\begin{split}
\largedot[1.4]{V}_2 &= -k_{\Theta}^{-1}k_{\bs{\omega}}\bs{\omega}_{\ts{e}}^T\bs{\omega}_{\ts{e}}+cm_{\ts{e}}\bs{\omega}_{\ts{e}}^T\bs{n}_{\ts{e}}+\frac{1}{2}cm_{\ts{e}}\bs{\omega}_{\ts{e}}^T\bs{\omega}_{\ts{e}}\\
&\hspace{3ex}-ck_{\Theta}\bs{n}_{\ts{e}}^T\bs{p}_{\ts{e},2} -ck_{\bs{\omega}}\bs{n}_{\ts{e}}^T\bs{\omega}_{\ts{e}}.
\end{split}
\label{EQ33}
\end{align}
This expression can be upper bounded as
\begin{align}
\begin{split}
\largedot[1.4]{V}_2 &\leq -k_{\Theta}^{-1}k_{\bs{\omega}}\|\bs{\omega}_{\ts{e}}\|_2^2+c\|\bs{\omega}_{\ts{e}}\|_2\|\bs{n}_{\ts{e}}\|_2+\frac{1}{2}c\|\bs{\omega}_{\ts{e}}\|_2^2\\
&\hspace{3ex}-ck_{\Theta}\|\bs{n}_{\ts{e}}\|_2^2 +ck_{\bs{\omega}}\|\bs{n}_{\ts{e}}\|_2\|\bs{\omega}_{\ts{e}}\|_2,
\end{split}
\label{EQ34}
\end{align}
because \mbox{$-\|\bs{n}_\ts{e}\|_2^2\geq -\|\bs{n}_{\ts{e}}\|_2\|\bs{p}_{\ts{e},2}\|_2$}, 
for any \mbox{$\Theta_{\ts{e}}\in \left[0, 2\pi\right)$}---the operational range selected by design. Last, using matrix manipulation rules, we obtain
\begin{align}
\begin{split}
\largedot[1.4]{V}_2 \leq 
-\bs{x}^T_{\ts{e}}
\left[
\begin{array}{cc}
ck_{\Theta} & -\frac{1}{2}c\left(k_{\bs{\omega}}+1\right) \\
\vspace{-2ex}
\\
-\frac{1}{2}c\left(k_{\bs{\omega}}+1\right) & ~~~k_{\Theta}^{-1}k_{\bs{\omega}}-\frac{1}{2}c
\end{array} 
\right]
\bs{x}_{\ts{e}},
\end{split}
\label{EQ35}
\end{align}
where \mbox{$\bs{x}_{\ts{e}} = \left[\|\bs{n}_{\ts{e}}\|_2~\|\bs{\omega}_{\ts{e}}\|_2\right]^T$}.
}
\end{document}